\documentclass{article}

\usepackage{graphicx} 
\usepackage{subfigure} 

\usepackage{natbib,natbibspacing}

\usepackage{times}
\usepackage{fix-cm}

\usepackage{algorithm}
\usepackage{algorithmic}

\usepackage{amsmath}
\usepackage{amssymb}
\usepackage{amsthm}

\newtheorem{lemma}{Lemma}[section]
\newtheorem{theorem}{Theorem}[section]
\newtheorem{corollary}{Corollary}[section]

\usepackage{hyperref}

\def\argmin{\mathop{\rm argmin}}

\usepackage[accepted]{icml2012}

\icmltitlerunning{Agnostic System Identification for Model-Based Reinforcement Learning}

\begin{document} 

\setlength\abovedisplayskip{5pt}
\setlength\belowdisplayskip{5pt}

\twocolumn[
\icmltitle{Agnostic System Identification\\ for Model-Based Reinforcement Learning}

\icmlauthor{St\'ephane Ross}{stephaneross@cmu.edu}
\icmladdress{Robotics Institute,
            Carnegie Mellon University, PA USA}
\icmlauthor{J. Andrew Bagnell}{dbagnell@ri.cmu.edu}
\icmladdress{Robotics Institute,
            Carnegie Mellon University, PA USA}

\icmlkeywords{system identification, reinforcement learning, online learning, agnostic learning}

\vskip 0.3in
]

\begin{abstract} 
A fundamental problem in control is to learn a model of a system from observations that is useful for controller synthesis. To provide good performance guarantees, existing methods must assume that the real system is in the class of models considered during learning. We present an iterative method with strong guarantees even in the agnostic case where the system is not in the class. In particular, we show that any \emph{no-regret online learning} algorithm can be used to obtain a near-optimal policy, provided some model achieves low training error and access to a good exploration distribution. Our approach applies to both discrete and continuous domains. We demonstrate its efficacy and scalability on a challenging helicopter domain from the literature.
\end{abstract} 

\section{Introduction}
\begin{figure}
\centering
\includegraphics[width=0.43\textwidth,trim=40 83 40 135,clip]{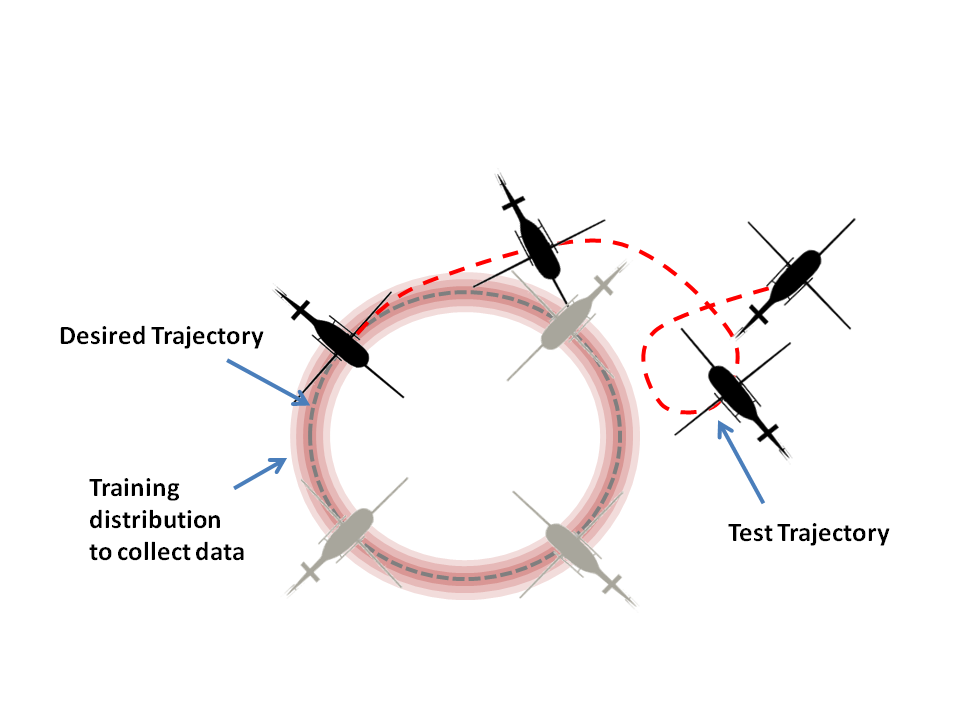}
\caption{Example train-test mismatch in a helicopter domain. Train: model is fit based on samples near the desired trajectory, e.g. from watching an expert. Test: learned policy ends up in new regions where model is bad, leading to poor control performance.\label{figMismatch}}
\vspace{-11pt}
\end{figure}

Model-based reinforcement learning (MBRL) and much of control rely on \emph{system identification}: building a model of a system from observations that is useful for controller synthesis. While often treated as a typical statistical learning problem, system identification presents different fundamental challenges as the executed controller and data generating process are inextricably intertwined. Naively attempting to estimate a controlled system can lead to a model that makes small error on a training set, but exhibits poor controller performance. This problem arises as the policy resulting from controller synthesis is often very different from the ``exploration'' policy used to collect data. While we might expect the model to make good predictions at states frequented by the exploration policy, the learned policy usually induces a different state distribution, where the model may poorly capture system behavior (Fig. \ref{figMismatch}).

This problem is fully appreciated in the system identification literature and has been attacked by considering ``open loop'' identification procedures and ``persistent excitation'' \citep{Ljung,Abbeel} that attempt to sufficiently ``cover'' the state-action space. Unfortunately, such methods rely on the strong assumption that the true system lies in the class of models considered: \textit{e.g.}, for continuous systems, they may require the true system to be modeled in a class of linear models. With this assumption, they ensure that eventually the correct model is learned-- \textit{e.g.}, by learning about every discrete state-action pair or all modes of the linear system-- to provide guarantees.

In this work, we provide algorithms for system identification and controller synthesis (\textit{i.e.} MBRL) that have strong performance guarantees with a weaker agnostic assumption that the system identification achieves statistically good prediction. We adopt a reduction-based analysis \citep{Beygelzimer05} that relates the learned policy's performance to prediction error during training. We begin by providing agnostic bounds for a simple generic ``batch'' algorithm that can represent many learning methods used in practice (\textit{e.g.}, building a model from open loop controls, watching an expert, or running a base policy we want to improve upon). Due to the mismatch in train/test distributions, uniform exploration is often the best option with this approach. Unfortunately, this makes the sample complexity and performance bounds scale with the size of the Markov Decision Process (MDP) (\textit{i.e.} state/action space). Next, we propose a simple iterative approach, closely related to online learning, with stronger guarantees that do not scale with the size of the MDP when given a good exploration distribution. The approach is very simple to implement and iterates between 1) collecting new data about the system by executing a good policy under the current model, as well as by sampling from a given exploration distribution, and 2) updating the model with that new data. 

This approach is inspired by a recent reduction of imitation learning to no-regret online learning \citep{DAgger} that addresses mismatch between train/test distributions. Our results can be interpreted as a \emph{reduction} of MBRL to no-regret online learning and optimal control, and show that any no-regret algorithm can be used in such a way to learn a policy with strong agnostic guarantees. This enables MBRL methods to match the strongest existing agnostic guarantees of model-free RL methods \citep{CPI, PSDP}.

We first introduce notation and related work. Then we present the batch method and our online learning approach with their agnostic guarantees (proofs are deferred to the supplementary material). Finally we demonstrate the efficacy of our approach on a challenging domain from the literature: learning to perform aerobatic maneuvers with a simulated helicopter \citep{Abbeel}.

\section{Background and Notation}
We assume the real system behaves according to some unknown MDP, represented by a set of states $S$ and actions $A$ (both potentially infinite and continuous), a transition function $T$, where $T_{sa}$ denotes the next state distribution if we do action $a$ in state $s$, and the initial state distribution $\mu$ at time 1. We assume the cost function $C : S \times A \rightarrow \mathbb{R}$ is known and seek to minimize the expected sum of discounted costs over an infinite horizon with discount $\gamma$.


For any policy $\pi$, let $\pi_s$ be the action distribution performed by $\pi$ in state $s$; $D^t_{\omega,\pi}$ the state-action distribution at time $t$ if we started in state distribution $\omega$ at time 1 and followed $\pi$; $D_{\omega,\pi} = (1-\gamma) \sum_{t=1}^\infty \gamma^{t-1} D^t_{\omega,\pi}$ the state-action distribution over the infinite horizon if we follow $\pi$, starting in $\omega$ at time 1; $V_\pi(s) = \mathbb{E}_{a \sim \pi_s, s' \sim T_{sa}}[C(s,a) + \gamma V_\pi(s')]$ the value function of $\pi$ (the expected sum of discounted costs of following $\pi$ starting in state $s$); $Q_{\pi}(s,a) = C(s,a) + \gamma \mathbb{E}_{s' \sim T_{sa}}[V_\pi(s')]$ the action-value function of $\pi$ (the expected sum of discounted costs of following $\pi$ after starting in $s$ and performing action $a$); and $J_{\omega}(\pi) = \mathbb{E}_{s \sim \omega}[V_\pi(s)] = \frac{1}{1-\gamma} \mathbb{E}_{(s,a) \sim D_{\omega,\pi}}[C(s,a)]$ the expected sum of discounted costs of following $\pi$ starting in  $\omega$. 

Our goal is to obtain a policy $\pi$ with small regret, \textit{i.e.} for any policy $\pi'$, $J_{\mu}(\pi) - J_{\mu}(\pi')$ is small. This is achieved indirectly by learning a model $\hat{T}$ of the system and solving for a (near-)optimal policy (under $\hat{T}$); \textit{e.g.}, using dynamic programming \citep{Puterman} or approximate methods \citep{FVI,Williams92}. For continuous systems, an important special case is linear models with quadratic cost functions, and potentially additive Gaussian noise, known as Linear Quadratic Regulators (LQR)\footnote{LQR is defined by 4 matrices $A$,$B$,$Q$,$R$ s.t. $x_{t+1} = A x_t + B u_t + \xi_t$, for $x_t$ and $u_t$ the state and action at time $t$, and $\xi_t \sim N(0,\Sigma)$ is (optional) Gaussian white noise, and the cost $C(x,u) = x^\top Q x + u^\top R u$ ($Q \succeq 0$, $R \succ 0$). The optimal policy is linear ($u = K x$) and the value function is quadratic ($x^\top V x$). LQR can be solved by dynamic programming on $V$ and $K$.} which can be solved exactly and efficiently. Non-linear systems with non-quadratic cost functions can also be solved approximately (local optima) using efficient iterative linearization techniques such as iLQR\citep{iLQR}.

{\bf Related Work:} In contrast with ``textbook'' system identification methods, in practice control engineers often proceed iteratively to build good models for controller synthesis. A first batch of data is collected to fit a model and obtain a controller, which is then tested in the real system. If performance is unsatisfactory, data collection is repeated with different sampling distributions to improve the model where needed, until control performance is satisfactory. By doing so, engineers can use feedback of the policies found during training to decide how to collect data and improve performance. Such methods are commonly used in practice and have demonstrated good performance in the work of \citet{Atkeson97,Abbeel}. In both works, the authors proceed by fitting a first model from state transitions observed during expert demonstrations of the task, and at following iterations, using the optimal policy under the current model to collect more data and fit a new model with all data seen so far. \citet{Abbeel} show this approach has good guarantees in non-agnostic settings (for finite MDPs or LQRs), in that it must find a policy that performs as well as the expert providing the initial demonstrations. Our method can be seen as making algorithmic this engineering practice, extending and generalizing the previous methods of \citet{Atkeson97,Abbeel}, and suggesting slight modifications that provide good guarantees even in \emph{agnostic} settings.

Similarly, the Dataset Aggregation (DAgger) algorithm of \citet{DAgger} uses a similar data aggregation procedure over iterations to obtain policies that mimic an expert well in imitation learning. The authors show that such a procedure can be interpreted as an online learning algorithm \citep{Hazan06, Kakade08}, more specifically, Follow-the-(Regularized)-Leader \citep{Hazan06}, and that using any no-regret online algorithm ensures good performance. Our approach can be seen as an extension of DAgger to MBRL settings.

Our approach leverages the way agnostic model-free RL algorithms perform exploration. Methods such as Conservative Policy Iteration (CPI) \citep{CPI} and Policy-Search by Dynamic Programming (PSDP) \citep{PSDP} learn a policy directly by updating policy parameters iteratively. For exploration, they assume access to a state exploration distribution $\nu$ that they can restart the system from and can guarantee finding a policy performing nearly as well as any policies inducing a state distribution (over a whole trajectory) close to $\nu$. Similarly, our approach uses a state-action exploration distribution to sample transitions and allows us to guarantee small regret against any policy with a state-action distribution close to this exploration distribution. If the exploration distribution is close to that of a near-optimal policy, then our approach guarantees near-optimal performance, provided a good model of data exists. This allows our model-based method to match the strongest agnostic guarantees of existing model-free methods. Good exploration distributions can often be obtained in practice; \textit{e.g.}, from human expert demonstrations, domain knowledge, or from a desired trajectory we would like the system to follow. Additionally, if we have a base policy we want to improve, it can be used to generate the exploration distribution -- with potentially additional random exploration in the actions.

\section{A Simple Batch Algorithm}
We now describe a simple algorithm, refered to as \textit{Batch}, that can be used to analyze many common approaches from the literature, \textit{e.g.}, learning from a generative model\footnote{With a generative model, we can set the system to any state, perform any action to obtain a sample transition.}, open loop excitation or by watching an expert \citep{Ljung}. 

Let $\mathcal{T}$ denote the class of transition models considered, and $\nu$ a state-action exploration distribution we can sample the system from. \textit{Batch} first executes in the real system $m$ state-action pairs sampled i.i.d. from $\nu$ to obtain $m$ sampled transitions. Then it finds the best model $\hat{T} \in \mathcal{T}$ of observed transitions, and solves (potentially approximately) the optimal control (OC) problem with $\hat{T}$ and known cost function $C$ to return a policy $\hat{\pi}$ for test execution. 



\subsection{Analysis}
Our reduction analysis seeks to answer the following question: if \textit{Batch} learns a model $\hat{T}$ with small error on training data, and solves the OC problem well, what guarantees does it provide on control performance of $\hat{\pi}$? Our results illustrate the drawbacks of a purely batch method due to the mismatch in train-test distribution.

We measure the quality of the OC problem's solution as follows. For any policy $\pi'$, let $\epsilon^{\pi'}_{\textrm{oc}} = \mathbb{E}_{s \sim \mu}[\hat{V}^{\hat{\pi}}(s) - \hat{V}^{\pi'}(s)]$ denote how much better $\pi'$ is compared to $\hat{\pi}$ on model $\hat{T}$ ($\hat{V}^{\hat{\pi}}$ and $\hat{V}^{\pi'}$ are the value functions of $\hat{\pi}$ and $\pi'$ under learned model $\hat{T}$ respectively). If $\hat{\pi}$ is an $\epsilon$-optimal policy on $\hat{T}$ within some class of policies $\Pi$, then $\epsilon^{\pi'}_{\textrm{oc}} \leq \epsilon$ for all $\pi' \in \Pi$. A natural measure of model error that arises from our analysis is in terms of $L_1$ distance between the predicted and true next state's distributions. That is, we define $\epsilon^{\textrm{L1}}_{\textrm{prd}} = \mathbb{E}_{(s,a) \sim \nu}[||T_{sa} - \hat{T}_{sa}||_1]$ the predictive error of $\hat{T}$, measured in $L_1$ distance, under the training distribution $\nu$. However, the $L_1$ distance cannot be evaluated or optimized from sampled transitions during training (we observe samples from $T_{sa}$ but not the distribution). Therefore we also provide our bounds in terms of other losses we can minimize from samples. This directly relates control performance to the model's training loss. A convenient loss is the KL divergence between $T_{sa}$ and $\hat{T}_{sa}$: $\epsilon^{\textrm{KL}}_{\textrm{prd}} = \mathbb{E}_{(s,a) \sim \nu,s' \sim T_{sa}}[\log(T_{sa}(s'))-\log(\hat{T}_{sa}(s'))]$. Minimizing KL corresponds to maximizing the log likelihood of the sampled transitions. This is convenient for common model classes, such as linear models (as in LQR), where it amounts to linear regression. For particular cases where $\mathcal{T}$ is a set of deterministic models and the real system has finitely many states, the predictive error can be measured via a classification loss at predicting the next state: $\epsilon^{\textrm{cls}}_{\textrm{prd}} = \mathbb{E}_{(s,a) \sim \nu,s' \sim T_{sa}}[\ell(\hat{T},s,a,s')]$, for $\ell$ the 0-1 loss of whether $\hat{T}$ predicts $s'$ for $(s,a)$, or any upper bound on the 0-1 loss, \textit{e.g.}, the multi-class hinge loss if $\mathcal{T}$ is a set of SVMs. In this case, model fitting is a supervised classification problem and the guarantee is directly related to the training classification loss. These are related as follows:

\begin{lemma} \label{lemPred}
$\epsilon^{\textrm{L1}}_{\textrm{prd}} \leq \sqrt{2\epsilon^{\textrm{KL}}_{\textrm{prd}}}$ and $\epsilon^{\textrm{L1}}_{\textrm{prd}} \leq 2\epsilon^{\textrm{cls}}_{\textrm{prd}}$. The latter holds with equality if $\ell$ is the 0-1 loss.
\end{lemma}

In general, we can use any loss minimizable from samples that upper bounds $\epsilon^{\textrm{L1}}_{\textrm{prd}}$ for models in the class. Our bounds are also related to the mismatch between the exploration distribution $\nu$ and distribution induced by executing another policy $\pi$ starting in $\mu$, denoted $c^{\pi}_{\nu} = \sup_{s,a} \frac{D_{\mu,\pi}(s,a)}{\nu(s,a)}$. We assume the costs $C(s,a) \in [C_{\min},C_{\max}]$ $\forall (s,a)$. Let $C_{\textrm{rng}} = C_{\max} - C_{\min}$ and $H = \frac{\gamma C_{\textrm{rng}}}{(1-\gamma)^2}$. $H$ is a scaling factor that relates model error to error in total cost predictions.

\begin{theorem} \label{thmBatch}
The policy $\hat{\pi}$ is s.t. for any policy $\pi'$:
\begin{displaymath}
J_{\mu}(\hat{\pi}) \leq J_{\mu}(\pi') + \epsilon^{\pi'}_{\textrm{oc}} + \frac{c^{\hat{\pi}}_\nu + c^{\pi'}_\nu}{2} H \epsilon^{\textrm{L1}}_{\textrm{prd}}
\end{displaymath}
This also holds as a function of $\epsilon^{\textrm{KL}}_{\textrm{prd}}$ or $\epsilon^{\textrm{cls}}_{\textrm{prd}}$ using Lem. \ref{lemPred}. 
\end{theorem}

This bound indicates that if \textit{Batch} solves the OC problem well and $\hat{T}$ has small enough error under the training distribution $\nu$, then it must find a good policy. Importantly, this bound is tight: \textit{i.e.} we can construct examples where it holds with equality (see supplementary material). More interestingly is what happens as we collect more data. If the fitting procedure is consistent (i.e. picks a model with minimal loss in the class asymptotically), then we can relate this guarantee to the capacity of the model class to achieve low error under the training distribution $\nu$. We denote the modeling error, measured in $L_1$ distance, as $\epsilon^{\textrm{L1}}_{\textrm{mdl}} = \inf_{T' \in \mathcal{T}} \mathbb{E}_{(s,a) \sim \nu}[||T_{sa} - T'_{sa}||_1]$. Similarly, define $\epsilon^{\textrm{KL}}_{\textrm{mdl}} = \inf_{T' \in \mathcal{T}} \mathbb{E}_{(s,a) \sim \nu,s' \sim T_{sa}}[\log(T_{sa}(s'))-\log(T'_{sa}(s'))]$ and $\epsilon^{\textrm{cls}}_{\textrm{mdl}} = \inf_{T' \in \mathcal{T}} \mathbb{E}_{(s,a) \sim \nu,s' \sim T_{sa}}[\ell(T',s,a,s')]$. These are all 0 in realizable settings, but generally non-zero in agnostic settings. After sampling $m$ transitions, the generalization error $\epsilon^{\textrm{L1}}_{\textrm{gen}}(m,\delta)$ bounds with high probability $1-\delta$ the quantity $\epsilon^{\textrm{L1}}_{\textrm{prd}}-\epsilon^{\textrm{L1}}_{\textrm{mdl}}$. Similarly, $\epsilon^{\textrm{KL}}_{\textrm{gen}}(m,\delta)$ and $\epsilon^{\textrm{cls}}_{\textrm{gen}}(m,\delta)$ denote the generalization error for the KL and classification loss respectively. $\epsilon^{\textrm{cls}}_{\textrm{gen}}(m,\delta)$ can be related to the VC dimension (or multi-class equivalent) in finite MDPs.

\begin{corollary} \label{corBatch}
After observing $m$ transitions, with probability at least $1-\delta$, for any policy $\pi'$:
\begin{displaymath}
J_{\mu}(\hat{\pi}) \leq J_{\mu}(\pi') + \epsilon^{\pi'}_{\textrm{oc}} + \frac{c^{\hat{\pi}}_\nu + c^{\pi'}_\nu}{2} H [\epsilon^{\textrm{L1}}_{\textrm{mdl}} + \epsilon^{\textrm{L1}}_{\textrm{gen}}(m,\delta)].
\end{displaymath}
This also holds as a function of $\epsilon^{\textrm{KL}}_{\textrm{mdl}}+\epsilon^{\textrm{KL}}_{\textrm{gen}}(m,\delta)$ (or $\epsilon^{\textrm{cls}}_{\textrm{mdl}}+\epsilon^{\textrm{cls}}_{\textrm{gen}}(m,\delta)$) using Lem. \ref{lemPred}. 
In addition, if the fitting procedure is consistent in terms of $L_1$ distance (or KL, classification loss), then $\epsilon^{\textrm{L1}}_{\textrm{gen}}(m,\delta) \rightarrow 0$ (or $\epsilon^{\textrm{KL}}_{\textrm{gen}}(m,\delta)\rightarrow 0$, $\epsilon^{\textrm{cls}}_{\textrm{gen}}(m,\delta)\rightarrow 0$) as $m \rightarrow \infty$ for any $\delta > 0$.
\end{corollary}

The generalization error typically scales with the complexity of the class $\mathcal{T}$ and goes to 0 at a rate of $O(\frac{1}{\sqrt{m}})$ ($\tilde{O}(\frac{1}{m})$ in ideal conditions). Given enough samples, the dominating factor limiting performance becomes the modeling error: \textit{i.e.} the term $\frac{c^{\hat{\pi}}_\nu + c^{\pi'}_\nu}{2} H \epsilon^{\textrm{L1}}_{\textrm{mdl}}$ (or equivalently $\frac{c^{\hat{\pi}}_\nu + c^{\pi'}_\nu}{2} H \sqrt{2\epsilon^{\textrm{KL}}_{\textrm{mdl}}}$ and $(c^{\hat{\pi}}_\nu + c^{\pi'}_\nu) H \epsilon^{\textrm{cls}}_{\textrm{mdl}}$) quantifies how performance degrades for agnostic settings. 

{\bf Drawback of Batch:} The two factors $c^{\hat{\pi}}_\nu$ and $c^{\pi'}_\nu$ are qualitatively different. $c^{\pi'}_\nu$ measures how well $\nu$ explores state-actions visited by the policy $\pi'$ we compare to. This factor is inevitable: we cannot hope to compete against policies that spend most of their time where we rarely explore. $c^{\hat{\pi}}_\nu$ measures the mismatch in train-test distribution. Its presence is the major drawback of \textit{Batch}. As $\hat{\pi}$ cannot be known in advance, we can only bound $c^{\hat{\pi}}_\nu$ by considering all policies we could learn: $\sup_{\pi \in \Pi} c^\pi_\nu$. This worst case is likely to be realized in practice: if $\nu$ rarely explores some state-action regions, the model could be bad for these and significantly underestimate their cost. The learned policy is thus encouraged to visit these low-cost regions where few data were collected. To minimize $\sup_{\pi \in \Pi} c^\pi_\nu$, the best $\nu$ for \textit{Batch} is often a uniform distribution, when possible. This introduces a dependency on the number of states and actions (or state-action space volume) (\textit{i.e.} $c^{\hat{\pi}}_\nu + c^{\pi'}_\nu$ is $O(|S||A|)$) multiplying the modeling error. Sampling from a uniform distribution often requires access to a generative model. If we only have access to a reset model\footnote{To sample transitions with a reset model, we can only simulate the system forward in time, or reset to a random initial state.} and a base policy $\pi_0$ inducing $\nu$ when executed in the system, then $c^{\hat{\pi}}_\nu$ could be arbitrarily large (\textit{e.g.}, if $\hat{\pi}$ goes to 0 probability states under $\pi_0$), and $\hat{\pi}$ arbitrarily worse than $\pi_0$. 

In the next section, we show that iterative learning methods can leverage feedback of the learned policies to obtain bounds that do {\bf not} depend on $c_\nu^{\hat{\pi}}$. This leads to better guarantees when we have a good exploration distribution $\nu$ (\textit{e.g.}, that of a near-optimal policy), or when we can only collect data via a reset model. This also leads to better performance in practice as shown in the experiments.

\section{No-Regret Methods for Agnostic MBRL}
Our extension of DAgger to the MBRL setting proceeds as follows. Starting from an initial model $\hat{T}^1 \in \mathcal{T}$, solve (approximately) the OC problem with $\hat{T}^1$ to obtain policy $\pi_1$. At each iteration $n$, collect data about the system by sampling state-action pairs from distribution $\rho_n = \frac{1}{2} \nu + \frac{1}{2} D_{\mu,\pi_n}$: \textit{i.e.} w.p. $\frac{1}{2}$, sample a transition occurring from an exploratory state-action pair drawn from $\nu$ and add it to dataset $\mathcal{D}$, otherwise, sample a state transition occurring from running the current policy $\pi_n$ starting in $\mu$, stopping the trajectory w.p. $1-\gamma$ at each step and adding the last transition to $\mathcal{D}$. The dataset $\mathcal{D}$ contains all transitions observed so far over all iterations. Once data is collected, find the best model $\hat{T}^{n+1} \in \mathcal{T}$ that minimizes an appropriate loss (\textit{e.g.} regularized negative log likelihood) on $\mathcal{D}$, and solve (approximately) the OC problem with $\hat{T}^{n+1}$ to obtain the next policy $\pi_{n+1}$. This is iterated for $N$ iterations. At test time, we could either find and use the policy with lowest expected total cost in the sequence $\pi_{1:N}$, or use the uniform ``mixture'' policy\footnote{At start of any trajectory, the mixture policy picks uniformly randomly a policy in $\pi_{1:N}$, and uses it for the whole trajectory.} over $\pi_{1:N}$. We guarantee good performance for both. The last policy $\pi_N$ often performs equally well, it has been trained with most data. Our experimental results confirm this intuition. In theory, $\pi_N$ has good guarantees when the distributions $D_{\mu,\pi_i}$ converge to a small region in the space of distributions as $i \rightarrow \infty$, but we do not guarantee this always occurs. 

{\bf Implementation with Off-the-Shelf Online Learner:} DAgger as described can be interpreted as using a \emph{Follow-The-(Regularized)-Leader} (FTRL) online algorithm to pick the sequence of models: at each iteration $n$ we pick the best (regularized) model $\hat{T}^{n}$ in hindsight under all samples seen so far. In general, DAgger can also be implemented using any no-regret online algorithm (see Algorithm~\ref{algDAggerMBRL}) to provide good guarantees. This is done as follows. When minimizing the negative log likelihood, the loss function of the online learning problem at iteration $i$ is: $L^{\textrm{KL}}_i(\hat{T}) = \mathbb{E}_{(s,a) \sim \rho_i, s' \sim T_{sa}} [-\log(\hat{T}_{sa}(s'))]$. This can be estimated from sampled state transitions at iteration $i$, and evaluated for any model $\hat{T}$. The online algorithm is applied on the sequence of loss $L^{\textrm{KL}}_{1:N}$ to obtain a sequence of models $\hat{T}^{1:N}$ over the iterations. As before, each model $\hat{T}^i$ is solved to obtain the next policy $\pi_i$. By doing so, the online algorithm effectively runs over mini-batches of data collected at each iteration to update the model, and each mini-batch comes from a different distribution that changes as we update the policy. Similarly, in a finite MDP with a deterministic model class $\mathcal{T}$, we can minimize the 0-1 loss instead (or any upper bound such as hinge loss) where the loss at iteration $i$ is: $L^{\textrm{cls}}_i(\hat{T}) = \mathbb{E}_{(s,a) \sim \rho_i, s' \sim T_{sa}} [\ell(\hat{T},s,a,s')]$, for $\ell$ the particular classification loss. This corresponds to an online classification problem. For many model classes, the negative log likelihood and convex upper bounds on the 0-1 loss (such as hinge loss) lead to convex online learning problems, for which no-regret algorithms exist (\textit{e.g.}, gradient descent, FTRL). As shown below, if the sequence of models is no-regret, then performance can be related to the minimum KL divergence (or classification loss) achievable with model class $\mathcal{T}$ under the overall training distribution $\overline{\rho} = \frac{1}{N} \sum_{i=1}^N \rho_i$ (\textit{i.e.} a quantity akin to $\epsilon^{\textrm{KL}}_{\textrm{mdl}}$ or $\epsilon^{\textrm{cls}}_{\textrm{mdl}}$ for \textit{Batch}).
 
\begin{algorithm}
\begin{algorithmic}
\STATE \textbf{Input:} exploration distribution $\nu$, number of iterations $N$, number of samples per iteration $m$, cost function $C$, online learning procedure \textsc{OnlineLearner}, optimal control procedure \textsc{OCSolver}.
\vspace{5pt}
\STATE Get initial guess of model: $\hat{T}^1 \leftarrow \textsc{OnlineLearner}()$.
\STATE $\pi_1 \leftarrow \textsc{OCSolver}(\hat{T}^1,C)$.
\FOR{$n=2$ \textbf{to} $N$}
\FOR{$k=1$ \textbf{to} $m$}
\STATE With prob. $\frac{1}{2}$ sample $(s,a) \sim D_{\mu,\pi_{n-1}}$ using $\pi_{n-1}$, otherwise sample $(s,a) \sim \nu$. Obtain $s' \sim T_{sa}$
\STATE Add $(s,a,s')$ to $\mathcal{D}_{n-1}$.
\ENDFOR
\STATE Update model: $\hat{T}^n \leftarrow \textsc{OnlineLearner}(\mathcal{D}_{n-1})$.
\STATE $\pi_n \leftarrow \textsc{OCSolver}(\hat{T}^n,C)$.
\ENDFOR
\STATE \textbf{Return} the sequence of policies $\pi_{1:N}$.
\end{algorithmic}
\caption{DAgger algorithm for Agnostic MBRL.\label{algDAggerMBRL}}
\end{algorithm}

%
\subsection{Analysis}
Similar to our analysis of \textit{Batch}, we seek to answer the following: if there exists a low error model of training data, and we solve each OC problem well, what guarantees does DAgger provide on control performance? Our results show that by sampling data from the learned policies, DAgger provides guarantees that have no train-test mismatch factor, leading to improved performance. 

For any policy $\pi'$, define $\overline{\epsilon}^{\pi'}_{\textrm{oc}} = \frac{1}{N} \sum_{i=1}^N \mathbb{E}_{s \sim \mu}[\hat{V}_i(s) - \hat{V}^{\pi'}_i(s)]$, where $\hat{V}_i$ and $\hat{V}^{\pi'}_i$ are respectively the value function of $\pi_i$ and $\pi'$ under model $\hat{T}^i$. This measures how well we solved each OC problem on average over the iterations. For instance, if at each iteration $i$ we found an $\epsilon_i$-optimal policy within some class of policies $\Pi$ on learned model $\hat{T}^i$, then $\overline{\epsilon}^{\pi'}_{oc} \leq \frac{1}{N} \sum_{i=1}^N \epsilon_i$ for all $\pi' \in \Pi$. As in \textit{Batch}, the average predictive error of the models $\hat{T}^{1:N}$ can be measured in terms of the $L_1$ distance between the predicted and true next state distribution: $\overline{\epsilon}^{\textrm{L1}}_{\textrm{prd}} = \frac{1}{N} \sum_{i=1}^N \mathbb{E}_{(s,a) \sim \rho_i}[||\hat{T}^i_{sa} - T_{sa}||_1]$. However, as was discussed, the $L_1$ distance is not observed from samples which makes it hard to minimize. Instead we can define other measures which upper bounds this $L_1$ distance and can be minimized from samples, such as the KL divergence or classification loss: \textit{i.e.} $\overline{\epsilon}^{\textrm{KL}}_{\textrm{prd}} = \frac{1}{N} \sum_{i=1}^N \mathbb{E}_{(s,a) \sim \rho_i, s' \sim T_{sa}}[\log(T_{sa}(s))-\log(\hat{T}^i_{sa}(s'))]$ and $\overline{\epsilon}^{\textrm{cls}}_{\textrm{prd}} = \frac{1}{N} \sum_{i=1}^N \mathbb{E}_{(s,a) \sim \rho_i, s' \sim T_{sa}}[\ell(\hat{T}^i,s,a,s')]$. Now, given the sequence of policies $\pi_{1:N}$, let $\hat{\pi} = \argmin_{\pi \in \pi_{1:N}} J_{\mu}(\pi)$ be the best policy in the sequence and $\overline{\pi}$ the uniform mixture policy on the sequence.
\begin{lemma} \label{lemDAgger}
The policies $\pi_{1:N}$ are s.t. for any policy $\pi'$: 
\begin{displaymath}
J_{\mu}(\hat{\pi}) \leq J_{\mu}(\overline{\pi}) \leq J_{\mu}(\pi') + \overline{\epsilon}^{\pi'}_{\textrm{oc}} +  c^{\pi'}_{\nu} H \overline{\epsilon}^{\textrm{L1}}_{\textrm{prd}}
\end{displaymath}
This also holds as a function of $\overline{\epsilon}^{\textrm{KL}}_{\textrm{prd}}$ or $\overline{\epsilon}^{\textrm{cls}}_{\textrm{prd}}$ using Lem. \ref{lemPred}.
\end{lemma}

We note that $\overline{\epsilon}^{\textrm{KL}}_{\textrm{prd}} = \frac{1}{N}\sum_{i=1}^N L_i^{KL}(\hat{T}^i) - L_i^{KL}(T)$ and $\overline{\epsilon}^{\textrm{cls}}_{\textrm{prd}} = \frac{1}{N} \sum_{i=1}^N L_i^{cls}(\hat{T}^i)$. Using a no-regret algorithm on the sequence of losses $L_{1:N}^{KL}$ implies $\frac{1}{N}\sum_{i=1}^N L_i^{KL}(\hat{T}^i) \leq \inf_{T' \in \mathcal{T}}\frac{1}{N}\sum_{i=1}^N L_i^{KL}(T') + \overline{\epsilon}^{\textrm{KL}}_{\textrm{rgt}}$, for $\overline{\epsilon}^{\textrm{KL}}_{\textrm{rgt}}$ the average regret of the algorithm after $N$ iterations, s.t. $\overline{\epsilon}^{\textrm{KL}}_{\textrm{rgt}} \rightarrow 0$ as $N \rightarrow \infty$. This relates $\overline{\epsilon}^{\textrm{KL}}_{\textrm{prd}}$ to the modeling error of the class $\mathcal{T}$: $\overline{\epsilon}^{\textrm{KL}}_{\textrm{mdl}} = \inf_{T' \in \mathcal{T}} \mathbb{E}_{(s,a) \sim \overline{\rho}, s' \sim T_{sa}}[\log(T_{sa}(s))-\log(T'_{sa}(s'))]$, \textit{i.e.} $\overline{\epsilon}^{\textrm{KL}}_{\textrm{prd}} \leq \overline{\epsilon}^{\textrm{KL}}_{\textrm{mdl}} + \overline{\epsilon}^{\textrm{KL}}_{\textrm{rgt}}$, for $\overline{\epsilon}^{\textrm{KL}}_{\textrm{rgt}} \rightarrow 0$. Similarly define $\overline{\epsilon}^{\textrm{cls}}_{\textrm{mdl}} = \inf_{T' \in \mathcal{T}} \mathbb{E}_{(s,a) \sim \overline{\rho}, s' \sim T_{sa}}[\ell(T',s,a,s')]$ and by using a no-regret algorithm on $L_{1:N}^{cls}$, $\overline{\epsilon}^{\textrm{cls}}_{\textrm{prd}} \leq \overline{\epsilon}^{\textrm{cls}}_{\textrm{mdl}} + \overline{\epsilon}^{\textrm{cls}}_{\textrm{rgt}}$ for $\overline{\epsilon}^{\textrm{cls}}_{\textrm{rgt}} \rightarrow 0$. In some cases, even if the $L_1$ distance cannot be estimated from samples, statistical estimators can still be no-regret with high probability on the sequence of loss $L^{\textrm{L1}}_i(T') = \mathbb{E}_{(s,a) \sim \rho_i}[||T_{sa} - T'_{sa}||_1]$. This is the case in finite MDPs if we use the empirical estimator of $T$ based on data seen so far (see supplementary material). If we define $\overline{\epsilon}^{\textrm{L1}}_{\textrm{mdl}} = \inf_{T' \in \mathcal{T}} \mathbb{E}_{(s,a) \sim \overline{\rho}}[||T_{sa} - T'_{sa}||_1]$, this implies that $\overline{\epsilon}^{\textrm{L1}}_{\textrm{prd}} \leq \overline{\epsilon}^{\textrm{L1}}_{\textrm{mdl}} + \overline{\epsilon}^{\textrm{L1}}_{\textrm{rgt}}$, for $\overline{\epsilon}^{\textrm{L1}}_{\textrm{rgt}} \rightarrow 0$. Our main result follows:


\begin{theorem} \label{thmDAgger}
The policies $\pi_{1:N}$ are s.t. for any policy $\pi'$: 
\begin{displaymath}
J_{\mu}(\hat{\pi}) \leq J_{\mu}(\overline{\pi}) \leq J_{\mu}(\pi') + \overline{\epsilon}^{\pi'}_{\textrm{oc}} +  c^{\pi'}_{\nu} H [\overline{\epsilon}^{\textrm{L1}}_{\textrm{mdl}} + \overline{\epsilon}^{\textrm{L1}}_{\textrm{rgt}}]
\end{displaymath}
This also holds as a function of $\overline{\epsilon}^{\textrm{KL}}_{\textrm{mdl}} + \overline{\epsilon}^{\textrm{KL}}_{\textrm{rgt}}$ (or $\overline{\epsilon}^{\textrm{cls}}_{\textrm{mdl}} + \overline{\epsilon}^{\textrm{cls}}_{\textrm{rgt}}$) using Lem. \ref{lemPred}. If the fitting procedure is no-regret w.r.t the sequence of losses $L^{L1}_{1:N}$ (or $L^{KL}_{1:N}$, $L^{cls}_{1:N}$), then $\overline{\epsilon}^{\textrm{L1}}_{\textrm{rgt}} \rightarrow 0$ (or $\overline{\epsilon}^{\textrm{KL}}_{\textrm{rgt}} \rightarrow 0$,$\overline{\epsilon}^{\textrm{cls}}_{\textrm{rgt}} \rightarrow 0$) as $N \rightarrow \infty$.
\end{theorem}

Additionally, the performance of $\pi_N$ can be related to $\overline{\pi}$ if the distributions $D_{\mu,\pi_i}$ converge to a small region:
\begin{lemma} \label{lemDAggerLast}
If there exists a distribution $D^*$ and some $\epsilon_{\textrm{cnv}}^* \geq 0$ s.t. $\forall i$, $||D_{\mu,\pi_i} - D^*||_1 \leq \epsilon_{\textrm{cnv}}^* + \epsilon_{\textrm{cnv}}^i$ for some sequence $\{\epsilon_{\textrm{cnv}}^i\}^\infty_{i=1}$ that is $o(1)$, then $\pi_N$ is s.t.: 
\begin{displaymath}
J_{\mu}(\pi_N) \leq J_{\mu}(\overline{\pi}) + \frac{C_{\textrm{rng}}}{2(1-\gamma)} [2\epsilon_{\textrm{cnv}}^* + \epsilon_{\textrm{cnv}}^N + \frac{1}{N}\sum_{i=1}^N \epsilon_{\textrm{cnv}}^i]
\end{displaymath}
Thus: $\limsup_{N \rightarrow \infty} J_{\mu}(\pi_N) - J_{\mu}(\overline{\pi}) \leq \frac{C_{\textrm{rng}}}{1-\gamma} \epsilon_{\textrm{cnv}}^*$
\end{lemma}

Thm.~\ref{thmDAgger} illustrates how we can reduce the original MBRL problem to a \emph{no-regret online learning} problem on a particular sequence of loss functions. In general, no-regret algorithms have average regret of $O(\frac{1}{\sqrt{N}})$ ($\tilde{O}(\frac{1}{N})$ in ideal cases) such that the regret term goes to 0 at a similar rate to the generalization error term for \textit{Batch} in Cor.~\ref{corBatch}. Here, given enough iterations, the term $c^{\pi'}_{\nu} H \overline{\epsilon}^{\textrm{L1}}_{\textrm{mdl}}$ determines how performance degrades in the agnostic setting (or $c^{\pi'}_{\nu} H \sqrt{2\overline{\epsilon}^{\textrm{KL}}_{\textrm{mdl}}}$ or $2 c^{\pi'}_{\nu} H \overline{\epsilon}^{\textrm{cls}}_{\textrm{mdl}}$ if we use a no-regret algorithm on the sequence of KL or classification loss respectively). Unlike for \textit{Batch}, there is no dependence on $c^{\hat{\pi}}_\nu$, only on $c^{\pi'}_\nu$. Thus, if a low error model exists under training distribution $\overline{\rho}$, no-regret methods are guaranteed to learn policies that performs well compared to any policy $\pi'$ for which $c^{\pi'}_\nu$ is small. Hence, $\nu$ is ideally $D_{\mu,\pi}$ of a near-optimal policy $\pi$ (\textit{i.e.} explore where good policies go).

{\bf Finite Sample Analysis:} A remaining issue is that the current guarantees apply if we can evaluate the expected loss ($L^{\textrm{L1}}_i$, $L^{\textrm{KL}}_i$ or $L^{\textrm{cls}}_i$) exactly. This requires infinite samples at each iteration. If we run the no-regret algorithm on estimates of these loss functions, \textit{i.e.} loss on $m$ sampled transitions, we can still obtain good guarantees using martingale inequalities as in online-to-batch \citep{CesaBianchi04} techniques. The extra generalization error term is typically $O(\sqrt{\frac{\log(1/\delta)}{Nm}})$ with high probability $1-\delta$. While our focus is not on providing such finite sample bounds, we illustrate how these can be derived for two scenarios in the supplementary material. For instance, in finite MDPs with $|S|$ states and $|A|$ actions, if $\hat{T}^i$ is the empirical estimator of $T$ based on samples collected in the first $i-1$ iterations, then choosing $m=1$ and $N$ in $\tilde{O}(\frac{C_{\textrm{rng}}^2 |S|^2 |A| \log(1/\delta)}{\epsilon^2 (1-\gamma)^4})$ guarantees that w.p. $1-\delta$, for any policy $\pi'$:
\begin{displaymath}
J_\mu(\hat{\pi}) \leq J_\mu(\overline{\pi}) \leq J_\mu(\pi') + \overline{\epsilon}^{\pi'}_{\textrm{oc}} + O(c^{\pi'}_\nu \epsilon)
\end{displaymath}
Here, $\overline{\epsilon}_{\textrm{mdl}}$ does not appear as it is 0 (realizable case). Given a good state-action distribution $\nu$, the sample complexity to get a near-optimal policy is $\tilde{O}(\frac{C_{\textrm{rng}}^2 |S|^2 |A| \log(1/\delta)}{\epsilon^2 (1-\gamma)^4})$. This improves upon other state-of-the-art MBRL algorithms, such as $R_{\max}$, $\tilde{O}(\frac{C_{\textrm{rng}}^3 |S|^2 |A| \log(1/\delta)}{\epsilon^3 (1-\gamma)^6})$ \citep{Rmax} and a recent modification of $R_{\max}$, $\tilde{O}(\frac{C_{\textrm{rng}}^2 |S| |A| \log(1/\delta)}{\epsilon^2 (1-\gamma)^6})$ \citep{CsabaICML10} (when $|S| < \frac{1}{(1-\gamma)^2}$). Here, the dependency on $|S|^2|A|$ is due to the complexity of the class ($|S|^2|A|$ parameters). With simpler classes, it can have no dependency on the size of the MDP. In the supplementary material, we analyze a scenario where $\mathcal{T}$ is a set of kernel SVM (deterministic models) with RKHS norm bounded by $K$. Choosing $m=1$ and $N$ in $O(\frac{C_{\textrm{rng}}^2(K^2 + \log(1/\delta))}{\epsilon^2 (1-\gamma)^4})$ guarantees that w.p. $1-\delta$, for any policy $\pi'$:
\begin{displaymath}
J_\mu(\hat{\pi}) \leq J_\mu(\overline{\pi}) \leq J_\mu(\pi') + \overline{\epsilon}^{\pi'}_{\textrm{oc}} + 2 c^{\pi'}_\nu H \hat{\epsilon}^{\textrm{cls}}_{\textrm{mdl}} + O(c^{\pi'}_\nu \epsilon),
\end{displaymath}
\noindent for $\hat{\epsilon}^{\textrm{cls}}_{\textrm{mdl}}$ the multi-class hinge loss on the training set after $N$ iterations of the best SVM in hindsight. Thus, if we have a good exploration distribution and there exists a good model in $\mathcal{T}$ for predicting observed data, we obtain a near-optimal policy with sample complexity that depends only on the complexity of $\mathcal{T}$, not the size of the MDP. 

\section{Discussion}
We emphasize that we provide reduction-style guarantees. DAgger may sometimes fail to find good policies, \textit{e.g.}, when no model in the class achieves low error on the training data. However, DAgger guarantees that one of the following occur: either (1) we find good policies or (2) no models with low error on the aggregate dataset exist. If the latter occurs, we need a better model class. In contrast, \textit{Batch} can find models with low training error, but \emph{still} fail at obtaining a policy with good control performance, due to train/test mismatch. This occurs even in scenarios where DAgger finds good policies, as shown in the experiments.

DAgger needs to solve many OC problems. This can be computationally expensive, \textit{e.g.}, with non-linear or high-dimensional models. Many approximate methods can be used, \textit{e.g.}, policy gradient \citep{Williams92}, fitted value iteration \citep{FVI} or iLQR \citep{iLQR}. As the models often change only slightly from one iteration to the next, we can often run only a few iterations of dynamic programming/policy gradient from the last value function/policy to obtain a good policy for the current model. As long as we get good solutions on average, $\overline{\epsilon}^{\pi'}_{\textrm{oc}}$ remains small and does not hinder performance.

DAgger generalizes the approach of \citet{Atkeson97} and \citet{Abbeel} so that we can use any no-regret algorithm to update the model, as well as any exploration distribution. A key difference is that DAgger keeps an even balance between exploration data and data from running the learned policies. This is crucial to avoid settling on suboptimal performance in agnostic settings as the exploration data could be ignored if it occupies only a small fraction of the dataset, in favor of models with lower error on the data from the learned policies. With this modification, our main contribution is showing that such methods have good guarantees even in agnostic settings.

\section{Experiments on Helicopter Domain}
We demonstrate the efficacy of DAgger on a challenging problem: learning to perform aerobatic maneuvers with a simulated helicopter, using the simulator of \citet{Abbeel}, which has a continuous 21-dimensional state and 4-dimensional control space. We consider learning to 1) hover and 2) perform a ``nose-in funnel'' maneuver. We compare DAgger to \textit{Batch} with several choices for $\nu$: 1) $\nu_t$: adding small white Gaussian noise\footnote{Covariance of $0.0025 I$ for states  and $0.0001 I$ for actions.} to each state and action along the desired trajectory, 2) $\nu_e$: run an expert controller, and 3) $\nu_{en}$: run the expert controller with additional white Gaussian noise\footnote{Covariance of $0.0001 I$.} in the controls of the expert. The expert controller is obtained by linearizing the true model about the desired trajectory and solving the LQR (iLQR for the nose-in funnel). We also compare against Abbeel's algorithm, where the expert is only used at the first iteration. 

\begin{figure}[t!]
\centering
\includegraphics[width=0.465\textwidth,trim=10 160 15 180,clip]{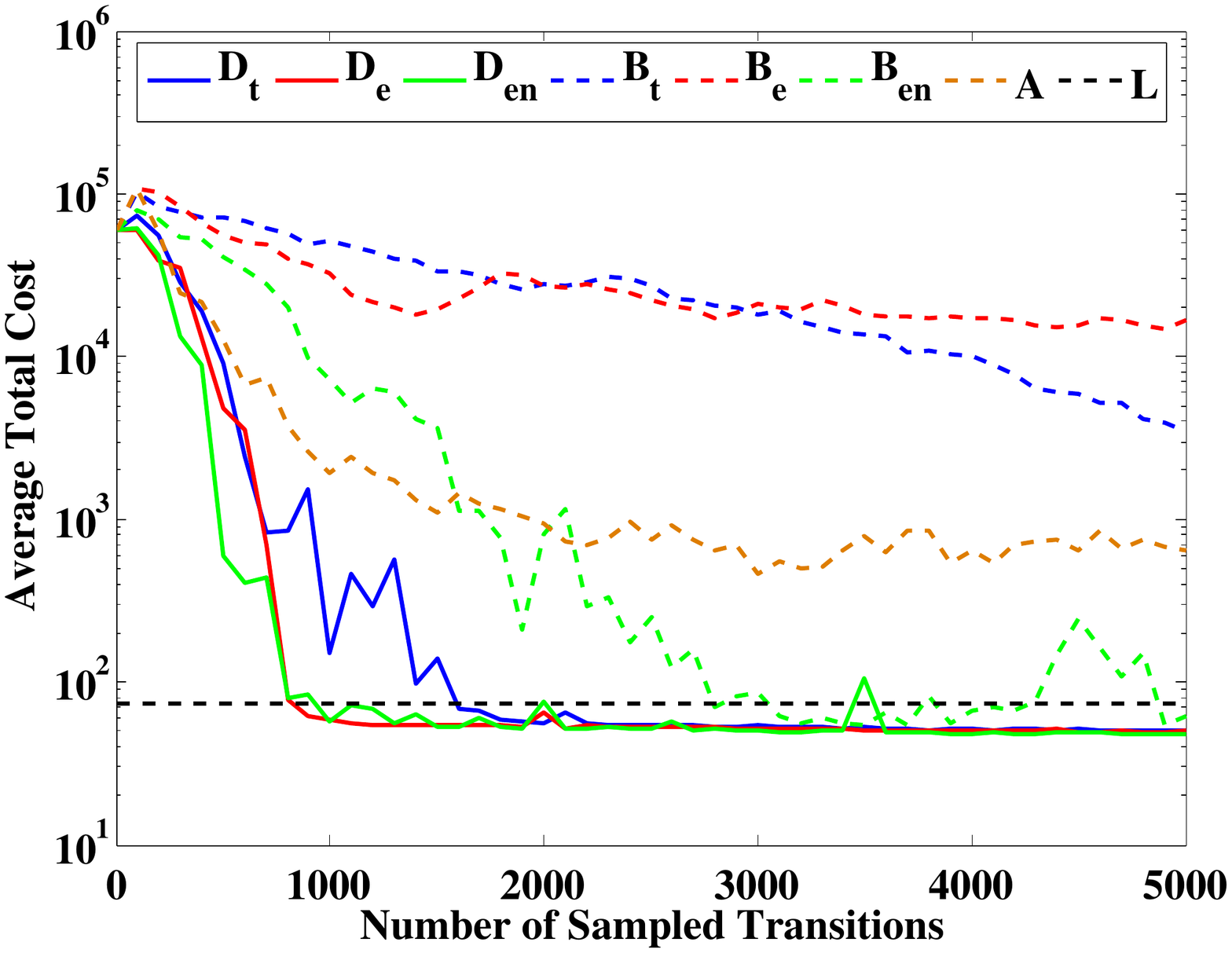}\\
\includegraphics[width=0.465\textwidth,trim=10 160 15 180,clip]{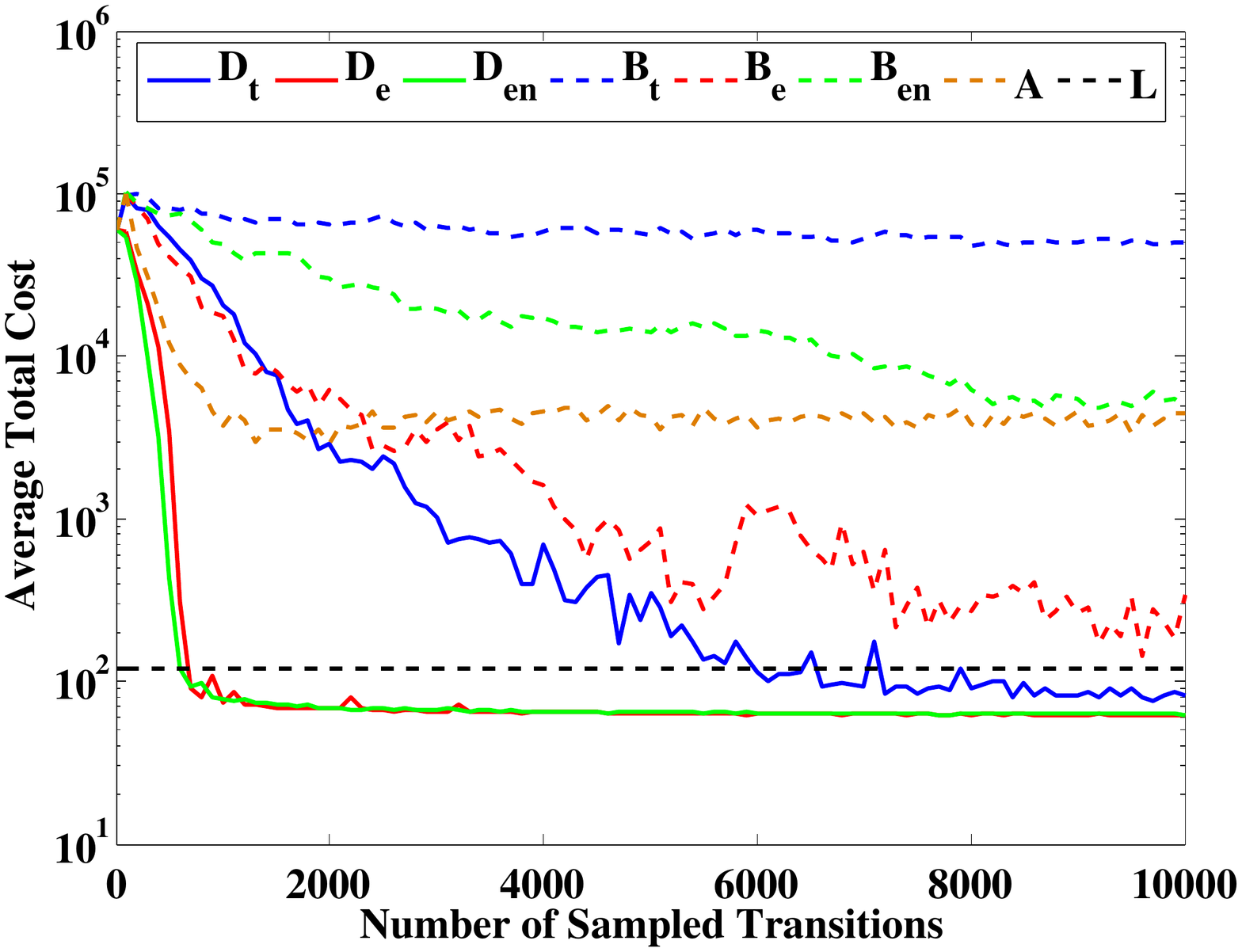} \\
\includegraphics[width=0.465\textwidth,trim=10 160 15 180,clip]{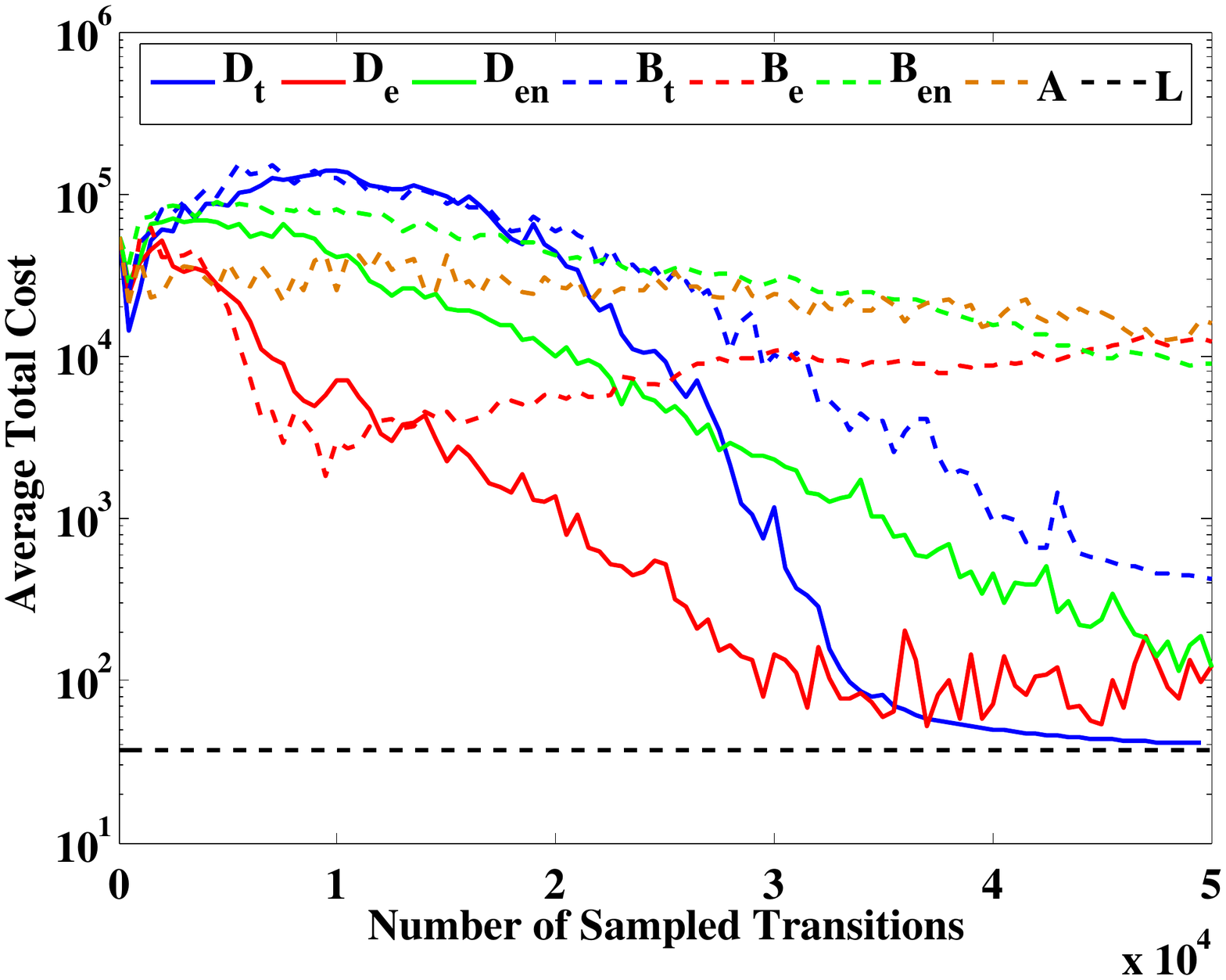} \\
\caption{Average total cost on test trajectories as a function of data collected so far, averaged over 20 repetitions of the experiments, each starting with a different random seed (all approaches use the same 20 seeds) From top to bottom: hover with no delay, hover with delay of 1, nose-in funnel. $D_t$, $D_e$ and $D_{en}$ denotes DAgger using exploration distribution $\nu_t$, $\nu_e$ and $\nu_{en}$ respectively, similarly $B_t$, $B_e$ and $B_{en}$ for the Batch algorithm, $A$ for Abbeel's algorithm, and $L$ for the linearized model's optimal controller. \label{figResults}}
\end{figure}

\textbf{Hover:} All approaches begin with an initial model $\Delta x_{t+1} = A \Delta x_t + B \Delta u_t$, for $\Delta x_t$ the difference between the current and hover state at time $t$, $\Delta u_t$ the delta controls at time $t$, $A$ is identity and $B$ adds the delta controls to the actual controls in $\Delta x_t$. We seek to learn offset matrices $A'$, $B'$ that minimizes $||\Delta x_{t+1} - [(A + A') \Delta x_t + (B + B') \Delta u_t]||_2$ on observed data\footnote{We also use a Frobenius norm regularizer on $A'$ and $B'$: $\min_{A',B'} \frac{1}{n} \sum_{i=1}^n ||\Delta x'_i - [(A + A') \Delta x_i + (B + B') \Delta u_i]||_2 + \frac{\lambda}{\sqrt{n}}(||A'||^2_F + ||B'||^2_F)$, for $\lambda = 10^{-3}$, $n$ the number of samples and $(\Delta x_i,\Delta u_i,\Delta x'_i)$ the $i^{th}$ transition in the dataset. During training we stop a trajectory if it becomes too far from the hover state, \textit{i.e.} if $||[\Delta x;\Delta u]||_2 > 5$ as this represents an event that would have to be recovered from. During testing, we run the trajectory until completion (400 timesteps of 0.05s, 20s total).}. We attempt to learn to hover in the presence of noise\footnote{White Gaussian noise with covariance $I$ on the forces and torques applied to the helicopter at each step.} and delay of 0 and 1. A delay of 1 introduces high-order dynamics that cannot be modeled with the current state. All methods sample 100 transitions per iteration and run for: 50 iterations when delay is 0; 100 iterations when delay is 1. Figure \ref{figResults} shows the test performance of each method after each iteration. In both cases, for any choice of $\nu$, DAgger outperforms \textit{Batch} significantly and converges to a good policy faster. DAgger is more robust to the choice of $\nu$, as it always obtains good performance given enough iterations, whereas \textit{Batch} obtains good performance with only one choice of $\nu$ in each case. Also, DAgger eventually learns a policy that outperforms the expert policy (L). As the expert policy is inevitably visiting states far from the hover state due to the large noise and delay (unknown to the expert), the linearized model is not as good at those states, leading to slightly suboptimal performance. Thus DAgger is learning a better linear model for the states visited by the learned policy which leads to better performance. Abbeel's algorithm improves the initial policy but reaches a plateau. This is due to lack of exploration (expert demonstrations) after the first iteration. While our objective is to show that DAgger outperforms other model-based approaches, we also compared against a model-free policy gradient method similar to CPI\footnote{Same as CPI, except gradient descent is done directly on deterministic linear controller. We solve a linear system to estimate the gradient from sample cost with perturbed  parameters.}. However, 100 samples per iteration were insufficient to get good gradient estimates and lead to only small improvement. Even with 500 samples per iteration, it could only reach an avg. total cost $\sim$15000 after 100 iterations.

{\textbf{Nose-In Funnel:}} This maneuver consists in rotating at a fixed speed and distance around an axis normal to the ground with the helicopter's nose pointing towards the axis of rotation (desired trajectory in Fig.~\ref{figMismatch}). We attempt to learn to perform 4 complete rotations of radius 5 in the presence of noise\footnote{Zero-mean spherical Gaussian with standard deviation 0.1 on the forces and torques applied to the helicopter at each step.} but no delay. We start each approach with a linearized model about the hover state and learn a time-varying linear model\footnote{For each time step $t$, we learn offset matrices $A'_t$, $B'_t$ such that $\Delta x_{t+1} = (A+A'_t)\Delta x_t + (B+B'_t) \Delta u_t + x^*_{t+1} - x^*_t$, for $x^*_t$ the desired state at time $t$ and $A,B$ the given hover model.}. All methods collect 500 samples per iteration over 100 iterations. Figure \ref{figResults} (bottom) shows the test performance after each iteration. With the initial model (0 data), the controller fails to produce the maneuver and performance is quite poor. Again, with any choice of $\nu$, DAgger outperforms \textit{Batch}, and unlike \textit{Batch}, it performs well with all choices of $\nu$. A video comparing qualitatively the learned maneuver with DAgger and \textit{Batch} is available on YouTube \citep{video}. Abbeel's method improves performance slightly but again suffers from lack of expert demonstrations after the first iteration. 

\section{Conclusion}
We presented a no-regret online learning approach to MBRL that has strong performance, both in theory and practice, even in agnostic settings. It is simple to implement, formalizes and makes algorithmic the engineering practice of iterating between controller synthesis and system identification, and can be applied to any control problem where approximately solving the OC problem is feasible. Additionally, its sample complexity scales with model class complexity, not the size of the MDP. To our knowledge, this is the first practical MBRL algorithm with agnostic guarantees. The only other agnostic MBRL approach we are aware of is a recent agnostic extension of $R_{\max}$ \citep{CsabaCOLT11} that is largely theoretical: it requires unknown quantities to run the algorithm (\textit{e.g.}, distance between the real system and the model class) and its sample complexity is exponential in the class complexity. 

\section*{Acknowledgements}
This work is supported by the ONR MURI grant N00014-09-1-1052, Reasoning in Reduced Information Spaces.

{\small
\bibliography{biblio}
\bibliographystyle{icml2012}}
\end{document}